\documentclass[11pt,a4paper]{article}
\usepackage[hyperref]{acl2020}
\usepackage{times}
\usepackage{latexsym}
\usepackage{multirow}
\usepackage{multicol}
\usepackage{booktabs}
\usepackage{graphicx}
\usepackage{amsmath}
\usepackage{bm}
\usepackage[utf8]{inputenc}
\usepackage{multirow}
\usepackage{cleveref}
\usepackage{xr-hyper}
\crefname{section}{§}{§§}
\Crefname{section}{§}{§§}

\usepackage{microtype}

\aclfinalcopy % Uncomment this line for the final submission
\def\aclpaperid{832} %  Enter the acl Paper ID here

\newcommand{\modelname}{{\usefont{T1}{ppl}{m}{n}LogicalFactChecker}}

\title{\modelname: Leveraging Logical Operations for Fact Checking with Graph Module Network}

\author{Wanjun Zhong$^1$\thanks{\ \ \ Work done while this author was an intern at Microsoft Research.} , Duyu Tang$^2$, Zhangyin Feng$^4$$^*$, Nan Duan$^2$, Ming Zhou$^2$\\
	\bf Ming Gong$^3$, Linjun Shou$^3$, Daxin Jiang$^3$, Jiahai Wang$^1$ and Jian Yin$^1$\\
	$^1$ The School of Data and Computer Science, Sun Yat-sen University.\\
	Guangdong Key Laboratory of Big Data Analysis and Processing, Guangzhou, P.R.China\\
	$^2$ Microsoft Research $^3$ Microsoft Search Technology Center Asia  $^4$ Harbin Institute of Technology \\
	{\tt \{zhongwj25@mail2,wangjiah@mail,issjyin@mail\}.sysu.edu.cn}\\
	{\tt \{dutang,nanduan,mingzhou,migon,lisho,djiang\}@microsoft.com}\\ 
	{\tt zyfeng@ir.hit.edu.cn}\\
}
\date{}

\begin{document}
\newcommand{\citett}[1]{\citeauthor{#1}~\shortcite{#1}}
	\maketitle
\begin{abstract}
		
Verifying the correctness of a textual statement requires not only semantic reasoning about the meaning of words, but also symbolic reasoning about logical operations like \textit{count}, \textit{superlative}, \textit{aggregation}, etc. In this work, we propose \modelname, a neural network approach capable of leveraging logical operations for fact checking. It achieves the state-of-the-art performance on TABFACT, a large-scale, benchmark dataset built for verifying a textual statement with semi-structured tables. This is achieved by a graph module network built upon the Transformer-based architecture. With a textual statement and a table as the input, \modelname \ automatically derives a program (a.k.a. logical form) of the statement in a semantic parsing manner. A heterogeneous graph is then constructed to capture not only the structures of the table and the program, but also the connections between inputs with different modalities. Such a graph reveals the related contexts of each word in the statement, the table and the program. The graph is used to obtain graph-enhanced contextual representations of words in Transformer-based architecture. After that, a program-driven module network is further introduced to exploit the hierarchical structure of the program, where semantic compositionality is dynamically modeled along the program structure with a set of function-specific modules. Ablation experiments suggest that both the heterogeneous graph and the module network are important to obtain strong results.
	\end{abstract}

	\section{Introduction}
	\begin{figure}[h]
    \centering
	\includegraphics[width=0.48\textwidth]{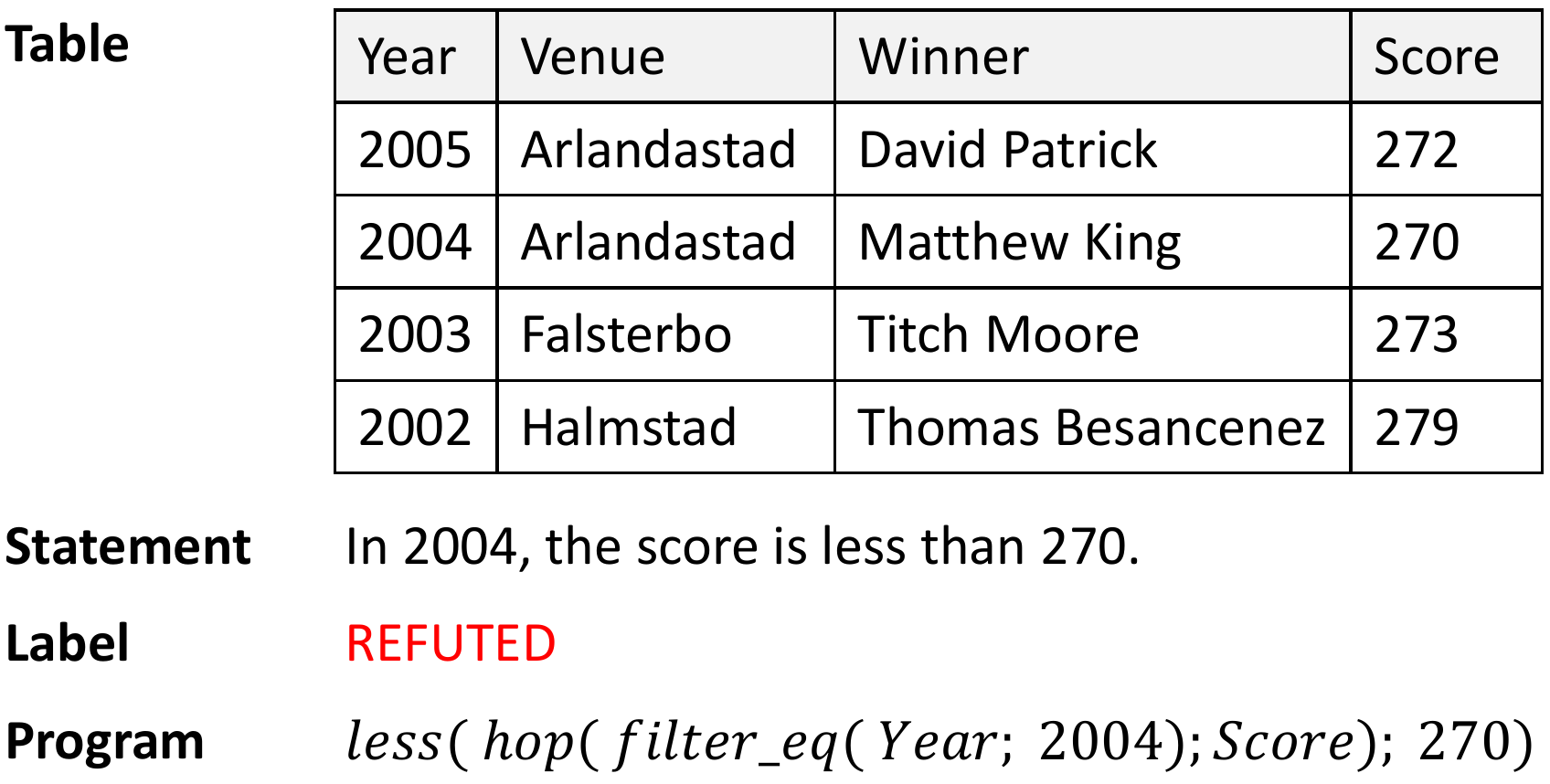}
	\caption{An example of table-based fact checking. Given a statement and a table as the input, the task is to predict the label. Program reflects the underlying meaning of the statement, which should be considered for fact checking.}
	\label{fig:task-example}
    \end{figure} 
    \begin{figure*}[t]    \centering
	\includegraphics[width=\textwidth]{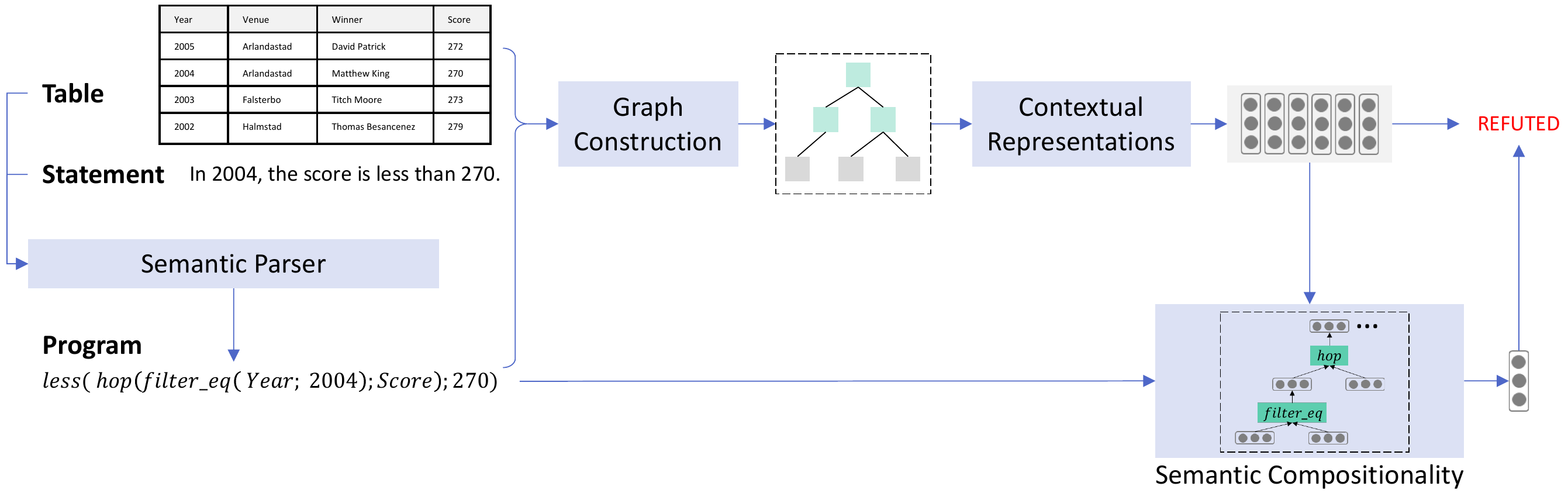}
	\caption{An overview of our approach \modelname. It includes a semantic parser to generate program (\cref{sec:program-generation}), a graph construction mechanism (\cref{sec:graph-construction}), a graph-based contextual representation learning for tokens (\cref{sec:graph-rep}) and 
% 		, we present how to utilize graph information for learning graph-enhanced contextual representations for tokens.
% 	Afterward, we introduce how we learn the 
	a semantic composition model over the program by neural module network (\cref{sec:comp-learning-module}).}
	\label{fig:pipeline-overview}
    \end{figure*}
	Fact checking for textual statements has emerged as an essential research topic recently because of the unprecedented amount of false news and rumors spreading through the internet \cite{thorne2018fever,2019TabFactA,goodrich2019assessing,nakamura2019r,kryscinski2019evaluating,vaibhav2019sentence}. 
	Online misinformation may manipulate people's opinions and lead to significant influence on essential social events like political elections \cite{faris2017partisanship}. 
% 	Most existing works only 
% 	Existing works on fact checking mainly focus on treating unstructured text as evidence. 
In this work, we study fact checking, with the goal of automatically assessing the truthfulness of a textual statement.

 The majority of previous studies in fact checking mainly focused on making better use of the meaning of words, while rarely considered symbolic reasoning about logical operations (such as ``\textit{count}'', ``\textit{superlative}'', ``\textit{aggregation}'').  However, modeling logical operations is an essential step 
 towards the modeling of complex reasoning and semantic compositionality.
%  for dealing with compositional and complex statements.
%  is appealing for 
%  	However, highly structured evidence (table, knowledge graph etc.) are also ubiquitous but have not been well exploited in the domain of fact checking. 
%  	Reasoning over the structured evidence is meaningful for applications like social network and financial analysis.
% 	In this paper, we explore the use of logical operations in fact checking.
% 	, with an application on table-based fact checking \cite{2019TabFactA},
% 	as a testbed to study the integration of logical operations in fact checking, which is 
% which is meaningful for applications like social networks and commercial management systems.
% 	, and financial analysis.
	% 	whose goal is to assess the veracity of a textual statement given a structured table as evidence.
	Figure \ref{fig:task-example} shows a motivating example for table-based fact checking, where the evidence used for verifying the statement comes from a semi-structured table.
% data as evidence
% 	that will be used throughout the paper. 
% 		Table-based fact checking is challenging because it requires systems to not only understand the semantic meaning of textual statement but also perform deep reasoning over the structure of table. 	Moreover, the symbolic operation is also essential in solving this task. 
We can see that correctly verifying the 
% 	For example, 
	statement 
	``\textit{In 2004, the score is less than 270}" requires a system 
	to not only discover the connections between tokens in the statement and the table, but more importantly understand the meaning of logical operations and how they interact 
% 	compose together
% recursive
	in a structural way to form a whole.
% 	produce the meaning representation.
Under this consideration, we use table-based fact checking as the testbed to 
% 	This motivates us to 
	investigate how to exploit logical operations in fact checking.
% 	, with an application on table-based fact checking.
	%semantically and structurally coherent way. 
% 	which reveal the full semantic and symbolic richness the 
%	execute condition operation and arithmetic (compare) operation for the final prediction.
%	From this example, we can see that reasoning ... has two major challenges.
%	The first is .... %(This is solved by your graph)
%	The second is .... %(this is solved by your module network and logical operations)
	
	% Moreover, the symbolic operation is also essential in solving this task. 
% 	Specifically, we work on TABFACT \cite{2019TabFactA}, which is a  large dataset for table-based fact checking. 
% 	In TABFACT, evidence comes from  semi-structured Wikipedia tables. 
	
	%TABFACT is challenging because it requires systems to not only understand the semantic meaning of textual statement but also perform deep reasoning over the structure of table. 
	% Moreover, the symbolic operation is also essential in solving this task. 
	% For example, statement ``in 2004, the score is less than 270" requires the system to execute condition operation and arithmetic (compare) operation for the final prediction.
% 	\par
	
	% are built based on the construction of program candidates and  
	% entity-linking and human-defined rules to construct LISP-like program candidates.
% 	\par
In this paper, We present \modelname, a neural network approach that leverages logical operations for fact checking when semi-structured tables are given as evidence.
% 	To address the aforementioned issues, 
% 	we present a transformer-based framework that concurrently utilizes the structural information in the table and the hierarchical information of symbolic operations in the program. 
% 	Our approach first construct the most consistent program with the
	Taking a statement and a table as the input, it first derives a program, also known as the logical form, in a semantic parsing manner \cite{liang2016learning}. 
	Then, our system builds a heterogeneous 
	graph to capture the connections among the statement, the table and the program.
	Such connections reflect the related context of each token in the graph, which are used to define attention masks in a Transformer-based \cite{vaswani2017attention} framework. The attention masks are used to
	learn graph-enhanced contextual representations of tokens\footnote{Here, tokens includes word pieces in the statement, table column names, table row names, table cells, and the program.}.
	We further develop a program-guided neural module network to capture the structural and compositional semantics of the program for semantic compositionality.
	%, inspired by 
	\cite{socher2013parsing,Andreas2015DeepCQ}.
	Graph nodes, whose representations are computed using the contextual representations of their constituents, are considered as arguments, and logical operations are considered as modules to recursively produce representations of higher level nodes along the program.
%	\cite{mitchell2010composition} 

	%, but the connection between
% 	To capture the connection 
% 	We further represent the context of the statement, the table and the constructed program as a graph retaining structural information of inputs.
% 	After obtaining the constructed graph, we present a transformer-based xxx model which applies a graph-based mask to control the related context within the inputs for learning graph-based contextual word representation.
% 	On top of this, we further present a graph-based reasoning model for learning the higher-level semantic compositionality.
	\par
	Experiments show that our system outperforms previous systems and achieves the state-of-the-art verification accuracy. The contributions of this paper can be summarized as follows:
	\begin{itemize}
		\item We propose \modelname, a graph-based neural module network, which utilizes logical operations for fact-checking.
		\item Our system achieves the state-of-the-art performance on TABFACT, a large-scale and benchmark dataset for table-based fact checking.  
		\item Experiments show that both the graph-enhanced contextual representation learning mechanism and the program-guided semantic compositionality learning mechanism improve the performance.
	\end{itemize}

	\section{Task Definition}
% 	Given a statement and a supporting semi-structured table as input, the task of table-based fact checking is to predict whether the statement is supported or refuted by the table.
We study the task of table-based fact checking in this paper. 
% 	In table-based fact checking, 
%	Given a textual statement and a supporting semi-structured table as evidence, the task is to assess the veracity of a statement. 
	This task is to assess the veracity of a statement when a table is given as evidence.
	Specifically, we evaluate our system on TABFACT \cite{2019TabFactA}, a large benchmark dataset for table-based fact checking. 
	With a given semi-structured table and a statement, systems are required to perform reasoning about the structure and content of the table and assess whether the statement is ``\textit{ENTAILED}" or ``\textit{REFUTED}" by the table.
	The official evaluation metric is the accuracy for the two-way classification (\textit{ENTAILED}/\textit{REFUTED}).
% 	\par
% 	TABFACT\cite{2019TabFactA} is a large-scale, benchmark dataset for table-based fact checking. 
	TABFACT consists of 118,439 statements and 16,621 tables from Wikipedia. More details about the dataset are given in Appendix A.
% 	given as evidence. 
% 	Therefore, we select TABFACT as a testbed for testing the ability of our system.
	
% 	As the example shown in Figure 1
% 	%  \ref{fig:task-example}
% 	, verification of a statement requires two forms of reasoning: 
% 	(1) linguistic reasoning: to verify the statements the systems are required to perform semantic-level understanding over the textual clues. For example, statement ``John J. Mcfall is unopposed during the re-election." requires the system to understand the semantic meaning of words.
% 	(2) logical reasoning: the verification also requires logical execution over the given table. For example, ``There are five candidates in total" requires the system to execute condition and arithmetic operation for the final prediction.
% 	\par
% 	In TABFACT, the official evaluation metric is the accuracy for the two-way classification (ENTAILED/REFUTED).
	
	\section{LogicalFactChecker: Methodology}
	
	In this section, we present 
	%  an overview of 
	%our approach LogicalFactChecker, which considers both semantic reasoning on the meaning of words and symbolic reasoning about logical operations for fact-checking.
	our approach \modelname, which simultaneously considers the meaning of words, inner structure of tables and programs, and logical operations for fact-checking. 
	One way to leverage program information is to use standard semantic parsing methods, where automatically generated programs are directly executed on tables to get results. However, TABFACT does not provide annotated programs. This puts the problem in a weak-supervised learning setting, which is one of the major challenges in the semantic parsing field. In this work, we use programs in a soft way that programs are represented with neural modules to guide the reasoning process between a textual statement and a table.
	%The goal of this model is to learn the semantic meaning of textual clues and synthesized programs in a soft way (compared to the discriminate execution) and predict the correct truthfulness.
	\par
	Figure \ref{fig:pipeline-overview} gives an overview of our approach.
	With a statement and a corresponding table, our system begins with program generation, which synthesizes a program. 
	Then, we build a heterogeneous graph for capturing the inner structure of the input.
%	tables and programs, and the connections between statements and them. 
	With the constructed graph, we incorporate a graph-based attention mask into the Transformer for learning graph-enhanced token representations. 
	Lastly, we learn the semantic compositionality by developing a program-guided neural module network and make the final prediction.

% 	LogicalFactChecker consists of following main components: 
% 	(1) \textbf{program generation module}; (2) \textbf{graph construction module}; (3) \textbf{graph-based representation learning module}; (4) \textbf{compositionality reasoning module} based on modular neural network.
	% 	Given a statement and a corresponding table as input, the \textbf{program generation module} first transfers the natural language statement to an executable program, which specifies both the logical operations used to represent the statement and the hierarchical relations between them. 
	% 	With generated program, the system builds a comprehensive graph based on the table, the statement and the program, and performs deep reasoning over the graph, finally states the veracity of the statement. 
	% 	The primary contribution of this paper is graph-based representation learning approach and logical reasoning approach in statement verification, which we will introduce in Section \ref{sec:statement-verification}. 
	% 	We will further describe the approach we apply for program generation in  \ref{sec:program-generation}.
	
	This section is organized as follows.
	We first describe the format of the program (\cref{section:program-representation}) for a more transparent illustration. 
	After that, the graph construction approach  (\cref{sec:graph-construction}) is presented first, followed by a graph-enhanced contextual representation learning mechanism (\cref{sec:graph-rep}). 
	Moreover, we introduce how to learn
% 		, we present how to utilize graph information for learning graph-enhanced contextual representations for tokens.
% 	Afterward, we introduce how we learn the 
	semantic compositionality over the program by neural module network (\cref{sec:comp-learning-module}). 
	At last, we describe how to synthesize programs by our semantic parsing model (\S \ref{sec:program-generation}).
	
	% \subsection{Graph construction}
	% 	Give a claim, a table and a 
	
	%  \section{Statement Verification Model}
	%  \label{sec:statement-verification}
	\subsection{Program Representation}
	\label{section:program-representation}
	Before presenting the technical details, we first describe the form of the program (also known as logical form) for clearer illustrations.
	\par
\begin{figure}[h]    \centering
	\includegraphics[width=0.5\textwidth]{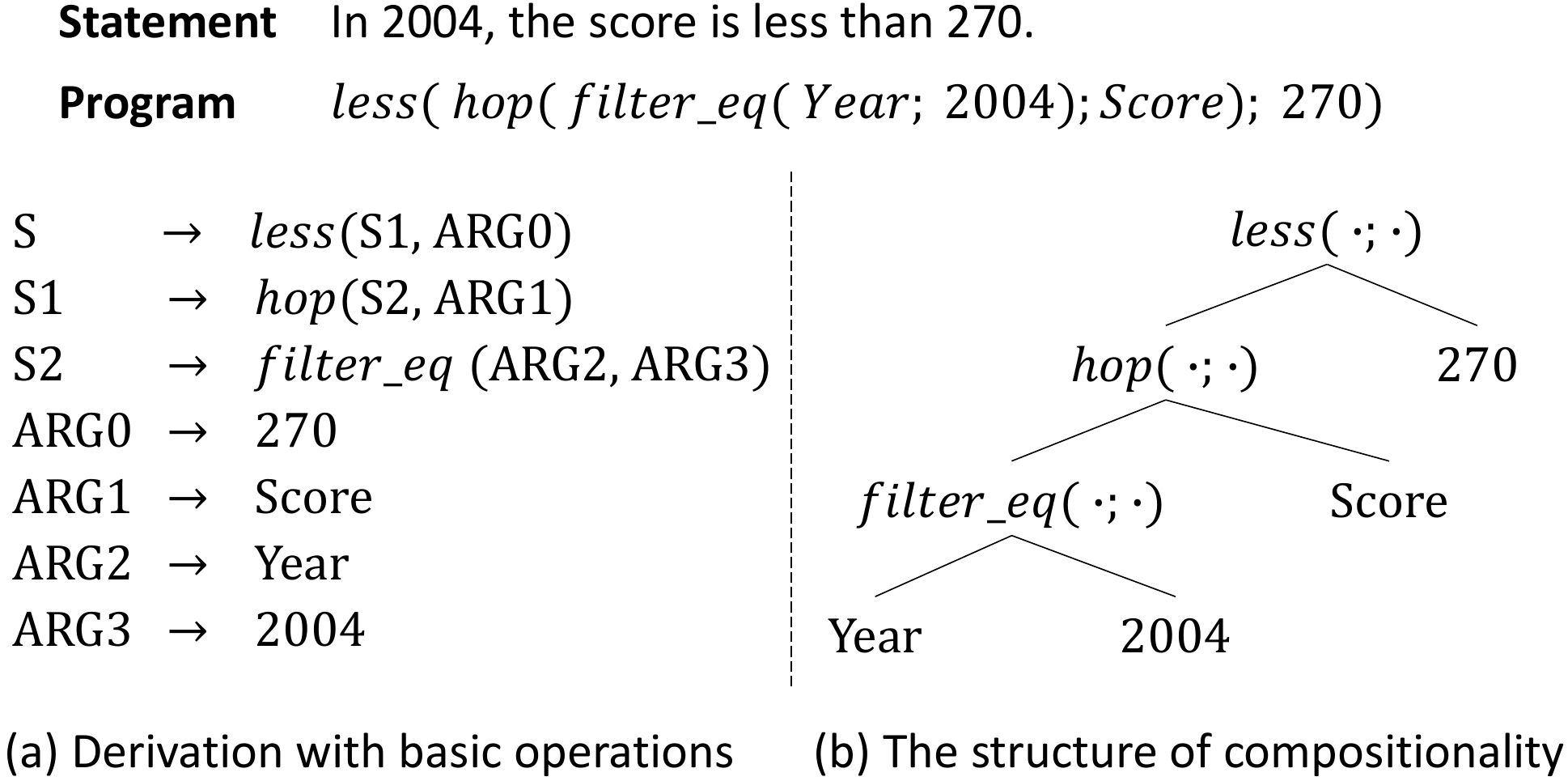}
	\caption{An example of a program with its semantic structure and derivation with basic logical operations.}
	\label{fig:program-representation}
    \end{figure}

 
	With a given natural language statement, we begin by synthesizing the corresponding semantic representation (LISP-like program here) using semantic parsing techniques.
	Following the notation defined by \citett{2019TabFactA}, 
	the functions (logical operations) formulating the programs come from a fixed set of over 50 functions, including ``\textit{count}" and ``\textit{argmax}", etc. 
%	 of the following 12 types: \{\textit{relationship}; \textit{selection}; \textit{filter}; \textit{choose}; \textit{verify}; \textit{query}; \textit{common}; \textit{different}; \textit{same}; \textit{and/or}; \textit{exist}\}.  
	The detailed description of the functions is given in Appendix C.
	Each function takes arguments of predefined types like string, number, bool or sub-table as input. 
	The programs have hierarchical structure because the functions can be nested.
	Figure \ref{fig:program-representation} shows an example of a statement and a generated program, accompanying with the derivation of the program and its semantic structure.   
	The details of the generation of a program for a textual statement are introduced in \cref{sec:program-generation}.
	
	\subsection{Graph Construction}
	\label{sec:graph-construction}
\begin{figure}[h]\centering
	\includegraphics[width=0.5\textwidth]{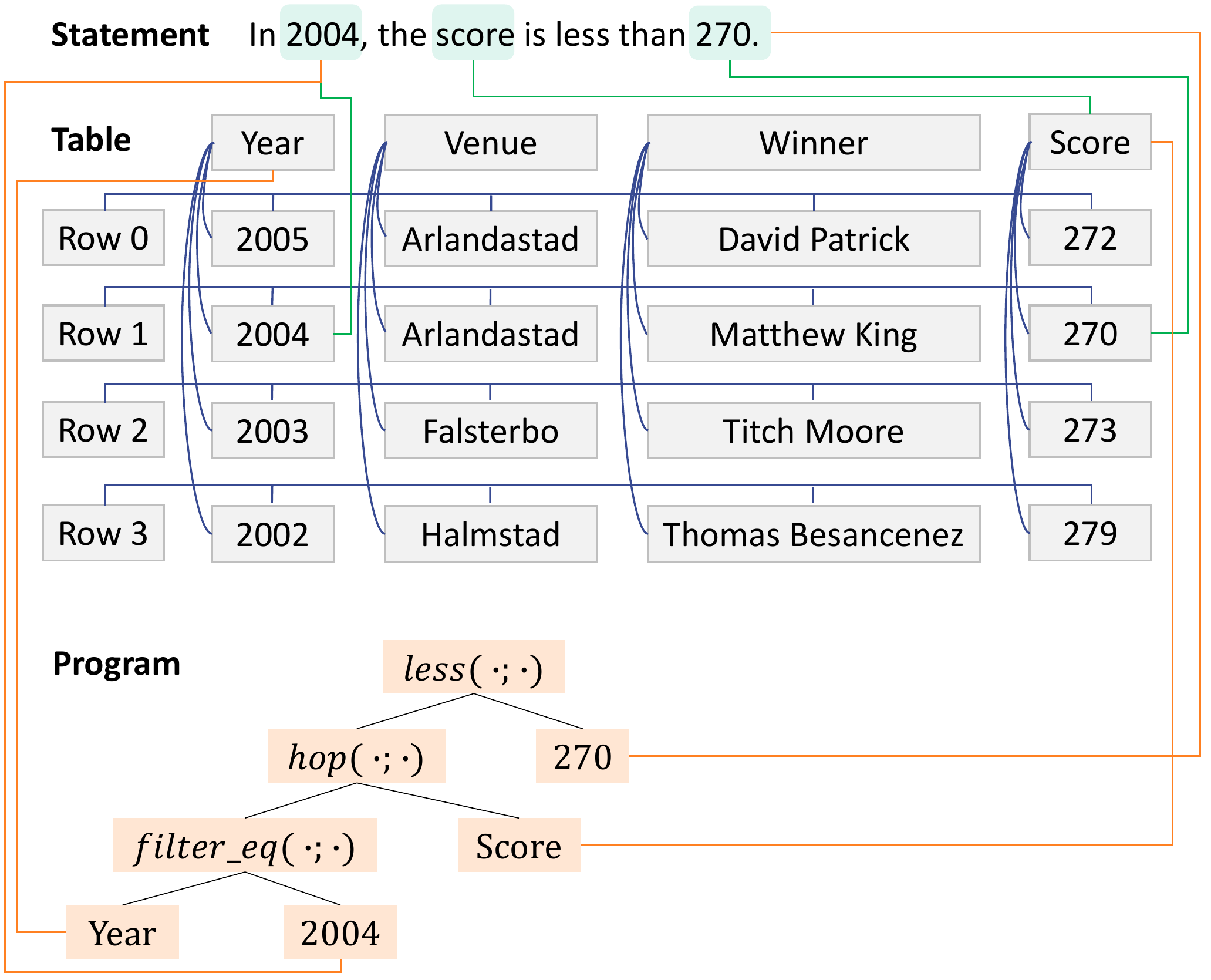}
	\caption{An example of the constructed graph.}
	\label{fig:constructed-graph}
    \end{figure}

	In this part, we introduce how to construct a graph to 
% 	reflect the 
% 	Our motivation for building a graph is that the graph can
	explicitly reveal the inner structure of programs and tables, and the connections among statements and them. 
% 	We construct the sub-graphs for tables, statements and programs in different ways and add connections to them.
% 	We use different ways to construct graphs for tables, statements, and programs. 
	Figure \ref{fig:constructed-graph} shows an example of the graph.
	%Our basic idea is :
	%partly inspired by the design of graph for table-based question answering \cite{}, we %define the contexts of each cell as its corresponding column name, other cells from %the same row in the table, and its occurrence in both statement and program. 
	Specifically, with a statement, a table and a program, our system operates in the following steps.
	\begin{itemize}
		\item 
		For a table, we define nodes as columns, cells, and rows, 
% 		Nodes from a table includes columns, cells, and rows. 
% 		We represent a table as a sub-graph by considering rows, columns, and cells as nodes, 
		which is partly inspired by the design of the graph for table-based question answering \cite{muller2019answering}. As shown in Figure \ref{fig:constructed-graph}, each cell is connected to its corresponding column node and row node. Cell nodes in the same row are fully-connected to each other. 
		\item 
		Program is a naturally structural representation consisting of functions and arguments.
		% 		naturally a graph because the program is a structural, formal meaning representation defined by hierarchical functions and their arguments. 
		In the program, functions and arguments are represented as nodes, and they are hierarchically connected along the structure.
% 		the function-argument edge. 
		Each node is connected to its direct parents and children.
		Arguments are also linked to corresponding column names of the table. 
		\item 
		By default, in the statement, all tokens are the related context of each other, so they are connected.
%		By default, each token in the statement connects to all other tokens in the statement because they are all related context of it.
%		each token in the statement is connected to all other tokens because the related contexts of 
%		 based on the consideration that the contexts of each token includes all other tokens in the statement. 
		 To further leverage the connections from the statement to the table and the program, we 
		% 		o the connection to tabla and program. We 
		add links for tokens which are linked to cells or columns in the table, and legitimate arguments in the program. 
% 		For the statement, we employ entity-linking technique to find entities mentioned concurrently in the statement, the table and the program. 
% 		These aligned entities are added to the graph as nodes and they are fully-connected with edges. 
% 		To incorporate the information from the statement into the graph, we use the entity-linking technique to align identical entities mentioned concurrently in the statement, table, and program. These aligned entities are connected by edges.  
	\end{itemize}
	After these processes, the extracted graph not only maintains the inner-structure of tables and programs but also explores the connections among aligned entities mentioned in different contents. 
	
	\subsection{Graph-Enhanced Contextual Representations of Tokens}
	\label{sec:graph-rep}
	%\subsubsection{Self-Attention}
	%The self-attention sublayers defined in Transformer \cite{}
	%\subsubsection{Representation Learning}
	
	We describe how to utilize the graph structure for learning graph-enhanced contextual representations of tokens \footnote{In this work, tokens include word pieces in the statement, column names and row names and contents of cells in the table, and function names in the program}. 
	%\subsubsection{Graph-enhanced co Representation}
	A simple way to learn contextual representations is to concatenate all the contents\footnote{All the contents indicate texts in the concatenated sequence of the linearized table, the statement, and the sequence of the linearized program.} as a single string and use the original attention mask in Transformer, where 
% 	regard all the tokens as the contexts 
% 	feed it into Transformer. 
% 	The original attention mask in Transformer takes 
	all the tokens are regarded as the contexts for each token. 
	However, this simple way fails to capture the semantic structure revealed in the constructed graph.
	For example, according to Figure \ref{fig:constructed-graph}, the content ``\textit{2004}" exists in the statement, program and table. 
	These aligned entity nodes for ``\textit{2004}" should be more related with each other when our model calculate contextual representations.
% 	For example, according to Figure \ref{fig:constructed-graph}, the token from node ``2004" in the table should be the related context of the token of node ``2004" in the statement and the program, but their positions are far away from each other in the concatenated sequence.
	To address this problem, we use the graph structure to re-define the related contexts of each token for learning a graph-enhanced representation. 
	%An intuitive way to model the structural information is to define an $N N$
	\par
	Specifically,  we present a graph-based mask matrix for self-attention mechanism in Transformer. 
	The graph-based mask matrix $G$ is a 0-1 matrix of the shape $N \times N$, where $N$ denotes the total number of tokens in the sequence. 
	This graph-based mask matrix records which tokens are the related context of the current token. 
	$G_{ij}$ is assigned as $1$ if token $j$ is the related context of token $i$ in the graph and $0$ otherwise.
%	For each token, the related contexts contain the contents of relative nodes in the graph.
%	Especially, to capture the complete semantic meaning, the related contexts of each token in the statement further includes all the tokens from the full statement.
	\par
	Then, the constructed graph-based mask matrix will be feed into BERT \cite{devlin2018bert} for learning graph-enhanced contextual representations.
% 	In this work, we employ BERT \cite{devlin2018bert} as the backbone of our approach. 
	We use the graph-based mask to control the contexts that each token can attend in the self-attention mechanism of BERT during the encoding process.
	%In this way, we obtain the graph-guided mask matrix between words, which will be feed into the model as a input.
%	After fine-tuning the BERT model with incorporating the graph-based mask matrix into the self-attention mechanism, we can obtain the graph-enhanced contextual representations for each token. 
	BERT maps the input $\bm{x}$ of length $T$ into a sequence of hidden vectors as follows.
	\begin{equation}
	h(\bm{x}) = [h(\bm{x})_1,h(\bm{x})_2,\cdots,h(\bm{x})_T]
	\end{equation} 
	These representations are enhanced by the structure of the constructed graph. 
	
	%  the multi-head attention is operated over al

	\subsection{Semantic Compositionality with Neural Module Network}
	\label{sec:comp-learning-module}
	In the previous subsection, we describe how our system learns the graph-enhanced contextual representations of tokens. 
	The process mentioned above learns the token-level semantic interaction. 
	In this subsection, we make further improvement by learning logic-level semantics using program information.
%	learning the structural semantic compositionality of logical operations in programs. 
% 	As described in the previous subsection, our system learns the graph-enhanced contextual representations for tokens.
% 	Beyond learning token-level semantic interaction, the system is also required to capture the structural semantic compositionality of logical operations in programs. 
% 	Beyond learning token-level contextual representation, we further want to learn deeper semantic compositionality of the logical operations.
	Our motivation is to utilize the structures and logical operations of programs for learning logic-enhanced compositional semantics.
%	design a network that can learn the semantic compositionality of programs.
	Since the logical operations forming the programs come from a fixed set of functions, we design a modular and composable network, where each logical operation is represented as a tailored module and modules are composed along the program structure. 
%	into a program-guided module network.
%     Since different programs have different hierarchical topologies for functions and arguments, 
% 	there is no a ``universal and best" network that can model the reasoning process of all the programs. 
% 	Therefore, our motivation is to design a modular and composable networks for learning semantic compositionality over the hierarchical structure of the program. 
	\par
%	After obtaining the graph-enhanced contextual representation of each token (\cref{sec:graph-rep}),
	We first describe how we initialize the representation for each entity node in the graph (\cref{sec:entity-node}). 
    After that, we describe how to learn semantic compositionality based on the program, including the design of each neural module (\cref{sec:modules}) and how these modules are composed recursively along the structure of the program (\cref{sec:comp-learning}).
    % Then we describe how to guide the learning process of semantic compositionality by the graph to verify the claim by considering the logical operations.
%    To achieve this, we propose to employ a neural module network, where each function in the program is implemented with a tailored neural module (\cref{sec:modules}).
%    and the learned entity representations are tackled as the input of modules. 
%    Finally, the semantic compositionality is learned by composing all the modules
%    recursively along the structure of the program (\cref{sec:comp-learning}).
	
	\subsubsection{Entity Node Representation}
	\label{sec:entity-node}
	
% 	The compositionality learning module, built on top of BERT, 
% 	takes graph-enhanced contextual representations to initialize the representations of every entity nodes in the program-based graph.
    
    In a program, entity nodes denote a set of entities (such as ``\textit{David Patrick}") from input contexts while function nodes denote a set of logical operations (such as``\textit{filter\_equal}"), both of which 
% 	Each entity node in the program-based graph 
	may contain multiple words/word-pieces.
% 	in the contexts. 
	Therefore, 
	we take graph-enhanced contextual representations as mentioned in \S \ref{sec:graph-rep} to initialize the representations of entity nodes.
	Specifically, we initialize the representation $h_e$ of each entity node $e$ by averaging the projected hidden vectors of each words contained in $e$ as follows:
% 	corresponding position of $h(\bm{x})$:
	\begin{equation}
	\label{equ:ent-representation}
	h_e = \frac{1}{n} \sum_{i=0}^{n} relu(\bm{W_e}h(\bm{x})_{p^i_e})
	\end{equation}
	where $n$ denotes the total number of tokens in the span of entity $e$, $p^i_e$ denotes the position of the $i^{th}$ token, $\bm{W_e} \in \bm{R}^{F\times D} $ is a weight matrix, $F$ is the dimension of feature vectors of arguments, $D$ is the dimension of hidden vectors of BERT and $relu$ is the activation function.
	\subsubsection{Modules}
	\label{sec:modules}
	\par
	In this part, we present function-specific modules, which are used as the basic computational units for composing all the required configurations of module network structures.
% 	we specify a set of modules that can compose into all the required configurations of network structure.  
	\par
% 	Each module corresponds to a specific function, where the function comes from a fixed set of over 40 functions described before. 
	Inspired by the neural module network \cite{Andreas2015DeepCQ} and the recursive neural network \cite{socher2013parsing}, we implement each module with the same neural architecture but with different function-specific parameters.
% 	and 
% 		is a tailored network that have its independent parameters and 
All the modules are trained jointly. 
% Specifically, we define each module as \cite{lapata-2001-corpus} 
Each module corresponds to a specific function, where the function comes from a fixed set of over 50 functions described before. 
In a program, each logical operation has the format of $\text{FUNCTION}(ARG0,ARG1,...)$, where each function
% each $\text{FUNCTION}$ comes from a fixed set of over 40 functions described before and 
may have variable-length arguments. For example, the function $hop$ has 2 arguments while the function $count$ has 1 argument.
To handle variable-length arguments, we develop each module as follows. We first calculate the composition for each function-argument pair and then produce the overall representation via combining the representations of items.

The calculation for each function-argument pair is implemented as matrix-vector multiplication, where each function is represented as a matrix and each argument is represented as a vector.
% , and 
This is inspired by vector-based semantic composition \cite{mitchell2010composition}, which states that matrix-vector multiplication could be viewed as the matrix modifying the meaning of vector.
% After having the composed representation for each argument, we 
% To handle variable-length arguments, we 
% implement each module as follows, 
% based on the consideration that functions are used to modify the meaning of arguments, and the overall representation is the combination of each constituent.
% consider that the 
% the meaning of each argument is modified by the function, and the final output is the combination of all these .
% Specifically, we represent a function as a matrix, and an argument as a vector for semantic composition \cite{mitchell2010composition}, where.
% matrix-vector multiplication could be viewed as an operation for modifying the meaning of representation of arguments 
% implement an module as follows, based on the consideration that the 
% matrix-vector multiplication could be viewed as an operation for modifying the meaning of representation of arguments
% % is regarded as the operator to 
% % the output is a combination over  \cite{mitchell2010composition}
% each $ARG$ is either a concrete entity node or the output of an intermediate logic
% inspired by the success of its application in visual question answering \cite{Andreas2015DeepCQ}.
% 	The arguments of a function can determine the number and type of inputs of a corresponding module.
% 	The inputs of each module either come from the representations of specific entity nodes or come from the outputs from other modules.
% 	\par
	Specifically, the output $y_m$ of module $m$ is computed with the following formula:
	\begin{equation}
	\label{equ:module-calculation}
	y_m = \frac{1}{N^m} \sum_{i=0}^{N^m} \sigma(\bm{W_m}v_i + b_m)
	\end{equation}
	where $\bm{W_m} \in \bm{R}^{F\times F}$ is a weight matrix and $b_m$ is a bias vector for a specific module $m$. $N^m$ denotes the number of arguments of module $m$, and each $v_i\in\bm{R}^{F}$ is the feature vector representing the $i^{th}$ input. 
% 	$v_i$ either comes from the representation of entity nodes or comes from the output of module from the previous step. 
	$\sigma$ is the activation function.
	
	Under the aforementioned settings, modules can compose into a hierarchical network determined by the semantic structure of the parsed program.
    
	\subsubsection{Program-Guided Semantic Compositionality}
	\label{sec:comp-learning}
\begin{figure}[h]    \centering
	\includegraphics[width=0.5\textwidth]{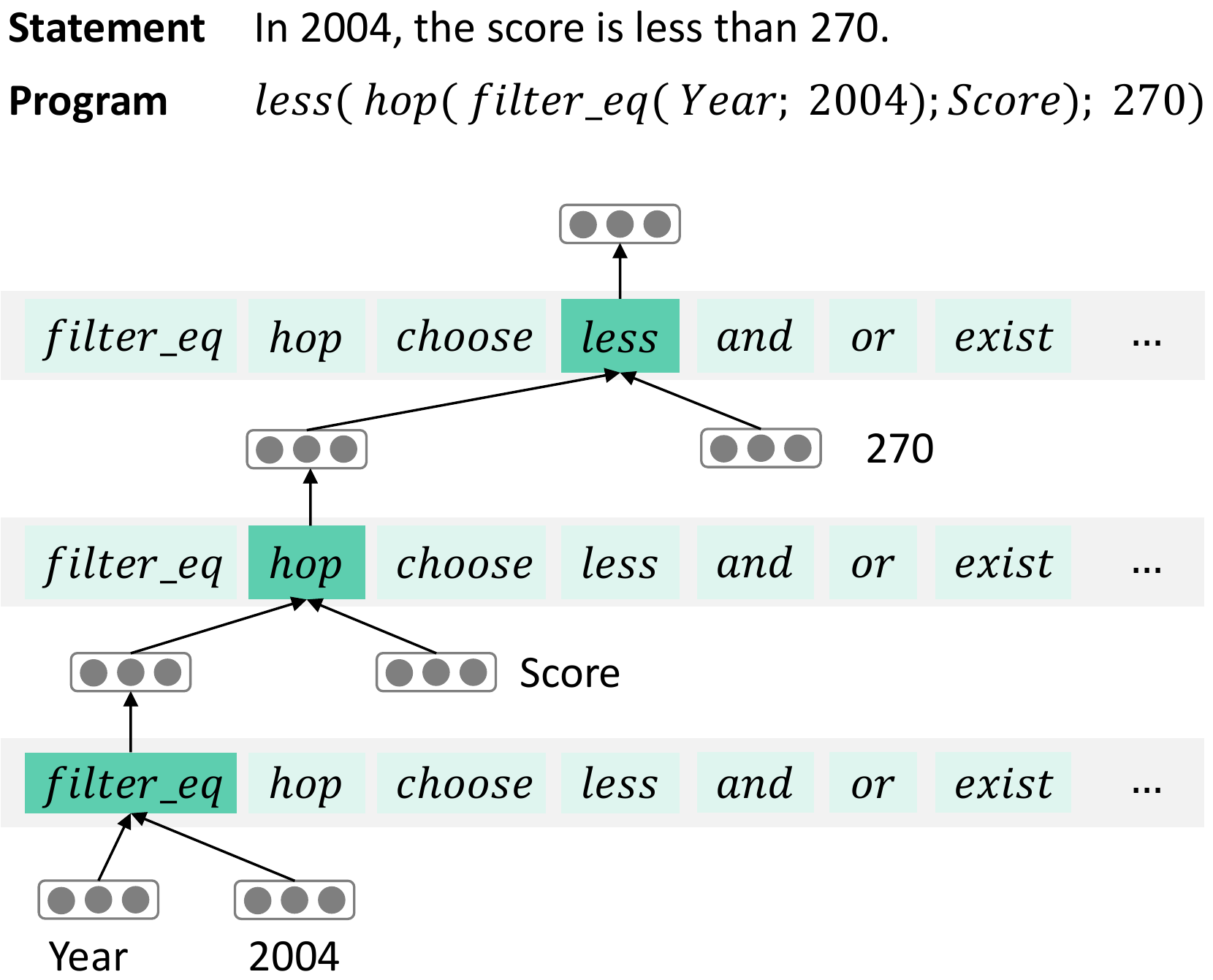}
	\caption{An example of neural module network.}
	\label{fig:compositionality}
    \end{figure}

	In this part, we introduce how to compose a program-guided neural module network based on the structure of programs and predefined modules.
	Taking the structure of the program and representations of all the entity nodes as the input, the composed neural module network learns the compositionality of the program for the final prediction. 
% 	Afterward, with composed network, we learn the semantic compositionality of the program for the final prediction.
    Figure \ref{fig:compositionality} shows an example of a composed network based on the structure of the program.
% 	The reasoning is performed recursively based on a bottom-up strategy. 
    % At first, we compute the representation of each entity node at the bottom of the program-based graph with Equation \ref{equ:ent-representation}. 
    % We obtain the calculated representations of entity nodes by process mentioned in \cref{sec:graph-rep}. 

    % each module takes the features of its arguments as inputs and operates on them to generate the output following a bottom-up order.
    
% 	Specifically, the output $y_m$ of module $m$ is computed with the following formula:
% 	\begin{equation}
% 	y_m = \frac{1}{N^m} \sum_{i=0}^{N^m} \sigma(\bm{W_m}v_i + b_m)
% 	\end{equation}
% 	where $\bm{W_m} \in \bm{R}^{F\times F}$ is a weight matrix and $b_m$ is a bias vector for a specific module $m$. $N^m$ denotes the number of arguments of module $m$, and each $v_i$ is the feature representing the $i^{th}$ argument. $v_i$ either comes from the representation of entity nodes or comes from the output of module from the previous step. $\sigma$ is the activation function.
    % At each step of compositionality learning, we select a module from a fixed set of parameterized modules defined in \cref{sec:modules}, and operate it to generate an intermidiate output.
    Along the structure of the program, each step of compositionality learning is to select a module from a fixed set of parameterized modules defined in \cref{sec:modules} and operate on it with Equation \ref{equ:module-calculation} to dynamically generate a higher-level representation.
    % Simulating the process of program execution, t
    The above process will be operated recursively until the output of the top-module is generated, which is denoted as $y_m^{top}$.
    % Along the structure of the program, the compositionality is learned by 
    % operating on modules and dynamically learning a higher level representation of program after each step.
% 	Simulating the process of program execution, the system operates calculation for each module with Equation \ref{equ:module-calculation} recursively until the final output of the top-module is generated, which is denoted as $y_m^t$.
	
	After that, we make the final prediction by feeding the combination of $y_m^{top}$ and the final hidden vector $h(\bm{x})_T$ from \cref{sec:graph-rep} through an MLP (Multi-layer Perceptron) layer.  
	The motivation of this operation is to retain the complete semantic meaning of the whole contexts because some linguistic cues are discarded during the synthesizing process of the program.

		\begin{table*}[ht]
\begin{tabular}{lccccc}
\hline
Model                                 & Val  & Test & \begin{tabular}[c]{@{}c@{}}Test\\ (simple)\end{tabular} & \begin{tabular}[c]{@{}c@{}}Test\\ (complex)\end{tabular} & Small Test  \\ \hline
Human Performance                     & -    & -    & -            & -             & 92.1       \\ \hline
Majority Guess                        & 50.7 & 50.4 & 50.8         & 50.0          & 50.3       \\
BERT classifier w/o Table             & 50.9 & 50.5 & 51.0         & 50.1          & 50.4       \\ \hline
Table-BERT (Horizontal-S+T-Concatenate) & 50.7 & 50.4 & 50.8         & 50.0          & 50.3       \\
Table-BERT (Vertical-S+T-Template)      & 56.7 & 56.2 & 59.8         & 55.0          & 56.2       \\
Table-BERT (Vertical-T+S-Template)      & 56.7 & 57.0 & 60.6         & 54.3          & 55.5       \\
Table-BERT (Horizontal-S+T-Template)    & 66.0 & 65.1 & 79.0         & 58.1          & 67.9       \\
Table-BERT (Horizontal-T+S-Template)    & 66.1 & 65.1 & 79.1         & 58.2          & 68.1       \\ \hline
LPA-Voting w/o Discriminator          & 57.7 & 58.2 & 68.5         & 53.2          & 61.5       \\
LPA-Weighted-Voting w/ Discriminator                 & 62.5 & 63.1 & 74.6         & 57.3          & 66.8       \\
LPA-Ranking w/ Discriminator             & 65.2 & 65.0 & 78.4         & 58.5          & 68.6       \\ \hline
% LogicalFactChecker (S+P+T)      & 70.8 & 71.3 &  85.3   &  64.5 & 73.8\\
LogicalFactChecker (program from LPA)      & 71.7 & 71.6 &  \textbf{85.5}   &  64.8 & 74.2\\
LogicalFactChecker (program from Seq2Action) & \textbf{71.8} & \textbf{71.7} &  85.4   &  \textbf{65.1} & \textbf{74.3} \\
 \hline
\end{tabular}
	\caption{Performance on TABFACT in terms of label accuracy (\%). The performances of Table-BERT and LPA are reported by \citet{2019TabFactA}. Our system is abbreviated as LogicalFactChecker, with program generated via our Sequence-to-Action model and baseline (i.e. LPA), respectively. T, S indicate the table, the statement and + means the order of concatenation.
	%T+S+P and S+P+T indicate the order of statement S and table T and program P. 
	In the linearization of tables, Horizontal (Vertical) refers to the horizontal (vertical) order for concatenating the cells. Concatenate (Template) means concatenating the cells directly (filling the cells into a template). In LPA settings, (Weighted) Voting means assigning each program with (score-weighted) equal weight to vote for the final result. Ranking means using the result generated by the top program ranked by the discriminator.}
	\label{tab:model-comparison}
	\end{table*}
	
	\subsection{Program Generation}
	\label{sec:program-generation}
	In this part, we describe our semantic parser
% 	ing based approaches 
	for synthesizing a program for a textual statement.
% 	semantic parsing. 
	We tackle the semantic parsing problem in a weakly-supervised setting \cite{berant2013semantic,liang2017neural,misra-EtAl:2018}, since the 
% 	problem, as described in \citet{liang2016neural}. 
% 	Since the 
	ground-truth program is not provided. 
	
% 	\par
	
	As shown in Figure \ref{fig:program-representation}, a program in TABFACT is structural and follows a grammar with over 50 functions. 
	To effectively capture the structure of the program and also generate legitimate programs following a grammar in the generation process, we develop a sequence-to-action approach, which is proven to be effective in solving many semantic parsing problems \cite{chen-etal-2018-sequence,iyer-etal-2018-mapping,guo2018dialog}.
	The basic idea is that the generation of a program tree is equivalent to the generation of a sequence of action, which is a traversal of the program tree following a particular order, like depth-first, left-to-right order.
% 	breadth-first.
% 	
% 	, each of which is 
	Specifically, our semantic parser works in a top-down manner in a sequence-to-sequence paradigm.
	The generation of a program follows an ASDL grammar \cite{yin2018tranx}, which is given in Appendix C.
	At each step in the generation phase, candidate tokens to be generated are only those legitimate according to the grammar.
% 	the generation of a token 
	Parent feeding \cite{yin2017syntactic} is used for directly passing information from parent actions.
	We further regard column names of the table as a part of the input \cite{zhong2017seq2sql} to generate column names as program arguments.

	We implement the approach with the LSTM-based recurrent network and Glove word vectors \cite{pennington2014glove} in this work, and the framework could be easily implemented with Transformer-based framework. 
	Following \citet{2019TabFactA}, we employ the label of veracity to guide the learning process of the semantic parser. 
	We also employ programs produced by LPA (Latent Program Algorithm) for comparison, which is provided by \citet{2019TabFactA}.

	In the training process, we train the semantic parser and the claim verification model separately. The training of semantic parser includes two steps: candidate search and sequence-to-action learning. 
	For candidate search, we closely follow LPA by first collecting a set of programs which could derive the correct label and then using the trigger words to reduce the number of spurious programs. For learning of the semantic parser, we use the standard way with back propagation, by treating each (claim, table, positive program) as a training instance.
% 	

% 	\cite{liang2016learning}
	
% 	We employ two methods for program generation. 
% 	The first method is 

% 	The second method is TransX, as described in \citet{yin2018tranx}. 
% 	The first step of these two mechanisms is latent program search. 
% 	Latent program search is built based on a predefined function set, including over 40 functions, in which each function takes arguments of specific types as input and executes on the table. The algorithm takes the linked entities and numbers from the statements and tables as input, and search for candidate programs that can consumes the inputs and generate the result of boolean value. Further details can be found in \citet{2019TabFactA}.
% 	Afterward, LPA trains a discriminator for semantic matching between statements and program candidates by taking all the programs with consistent label as positive instances and other programs as negative instances and minimizing the cross-entropy loss.
% 	\begin{figure}[h]    \centering
% 	\includegraphics[width=0.5\textwidth]{program-generation.pdf}
% 	\caption{An example from the table-based fact checking.}
% 	\label{fig:pipeline-overview}
%     \end{figure}

	\section{Experiments}
	
	We evaluate our system on TABFACT \cite{2019TabFactA}, a benchmark dataset for table-based fact checking. Each instance in TABFACT consists of a statement, a semi-structured Wikipedia table and a label (``ENTAILED" or ``REFUTED") indicates whether the statement is supported by the table or not. 
	The primary evaluation metric of TABFACT is label accuracy. The statistics of TABFACT are given in Appendix A. 
	Detailed hyper-parameters for model training are given in Appendix B for better reproducibility of experiments.
	
% 	\subsection{Baselines}
	We compare our system with following baselines, including the textual matching based baseline Table-BERT and semantic parsing based baseline LPA, both of which are developed by \citet{2019TabFactA}.
	\begin{itemize}
		\item Table-BERT tackles the problem as a matching problem. It takes the linearized table and the statement as the input and employs BERT to predict a binary class.
		\item Latent Program Algorithm (LPA) formulates the verification problem as a weakly supervised semantic parsing problem. With a given statement, it operates in two step: (1) latent program search for searching executable program candidates and (2) transformer-based discriminator selection for selecting the most consistent program. The final prediction is made by executing the selected program.  
	\end{itemize}
	
	\subsection{Model Comparison}
	% \usepackage{multirow}
	% \usepackage{multirow}

	In Table \ref{tab:model-comparison}, we compare our model (\modelname) with baselines on the development set and test set. 
	It is worth noting that complex test set and simple test set are partitioned based on its collecting channel, where the former involves higher-order logic and more complex semantic understanding.
	As shown in Table \ref{tab:model-comparison}, our model with programs generated by Sequence-to-Action model, significantly outperforms previous systems with 71.8\% label accuracy on the development set and 71.7\% on the test set, and achieves the state-of-the-art performance on the TABFACT dataset.
% 	is also the 
% 	Our approach 

% 	Our system outperforms both of the textual matching based baseline Table-BERT and the semantic-parsing based baseline LPA, which reflects that our model is better at modeling the inner structure of contexts and semantic compositionality of the logical operations.
	
	\subsection{Ablation Study}
	% \usepackage{multirow}
	We conduct ablation studies to evaluate the effectiveness of different components in our model. 
% 	Table \ref{tab:ablation-study} reports the performance on development set and test set after removing different components, including
\begin{table}[h]
		\begin{tabular}{l|ll}
			\hline
			\multirow{2}{*}{Model}  & \multicolumn{2}{l}{Label Acc. (\%)} \\ \cline{2-3} 
			& Val         & Test        \\ \hline
			LogicalFactChecker      & 71.83            & 71.69           \\
			-w/o Graph Mask         & 70.06             & 70.13            \\
			-w/o Compositionality   & 69.62            & 69.61           \\\hline
			%-w/o both above mechanisms & 66.10             & 66.06            \\ \hline
		\end{tabular}
	\caption{Ablation studies on the development set and the test set.}
	\label{tab:ablation-study}
	\end{table}
	
    As shown in Table \ref{tab:ablation-study}, we evaluate \modelname\ under following settings: 
	(1) removing the graph-based mask described in \cref{sec:graph-rep} (the first row); (2) removing the program-guided compositionality learning mechanism described in \cref{sec:comp-learning-module} (the second row).
	%(3) removing both above mechanisms (the last row). 
	%Specifically, the setting in the last row is that we simply concatenate the linearized table, the statement and the program into a single string and feed it into BERT for classification without considering any graph structure. 
	
% 	The BERT baseline (last row in Table \ref{tab:ablation-study}) take the linearized table, statement and linearized program as input, and directly take the last hidden vector from original BERT through an MLP layer for final prediction. This baseline utilize no graph information for verification.
	
	\par
	Table \ref{tab:ablation-study} shows that, eliminating the graph-based mask drops the accuracy by 
% 	1.77\% on development set and 
	1.56\% on test set.
	% 	which reflects that graph-enhanced representation learning module can help the model to better capture the relative information of linked nodes based on graph structure.
	Removing the program-guided compositionality learning mechanism drops the accuracy by 
% 	2.21\% on development set and 
	2.08\% on test set, which reflects that the neural module network plays a more important role in our approach.
% 	
% 	
% 	can reflect that this module can help to learn the semantic compositionality of logical operations. 
	%Removing both mechanisms leads to a further drop in the performance, which 
	This observation verifies that both mechanisms are beneficial for our task.
% 	compared to the BERT baseline, incorporating the graph-based representation learning module and the compositionality learning module improve the accuracy by xxx \%. 
% 	Eliminating the graph-enhanced representation learning module drop the label accuracy by xxx\%. 
    
% 	The experiments reflects that graph-enhanced representation learning module can help the model to better capture the relative information of linked nodes based on graph structure. 

	\subsection{Case Study}
\begin{figure}[h]    \centering
	\includegraphics[width=0.47\textwidth]{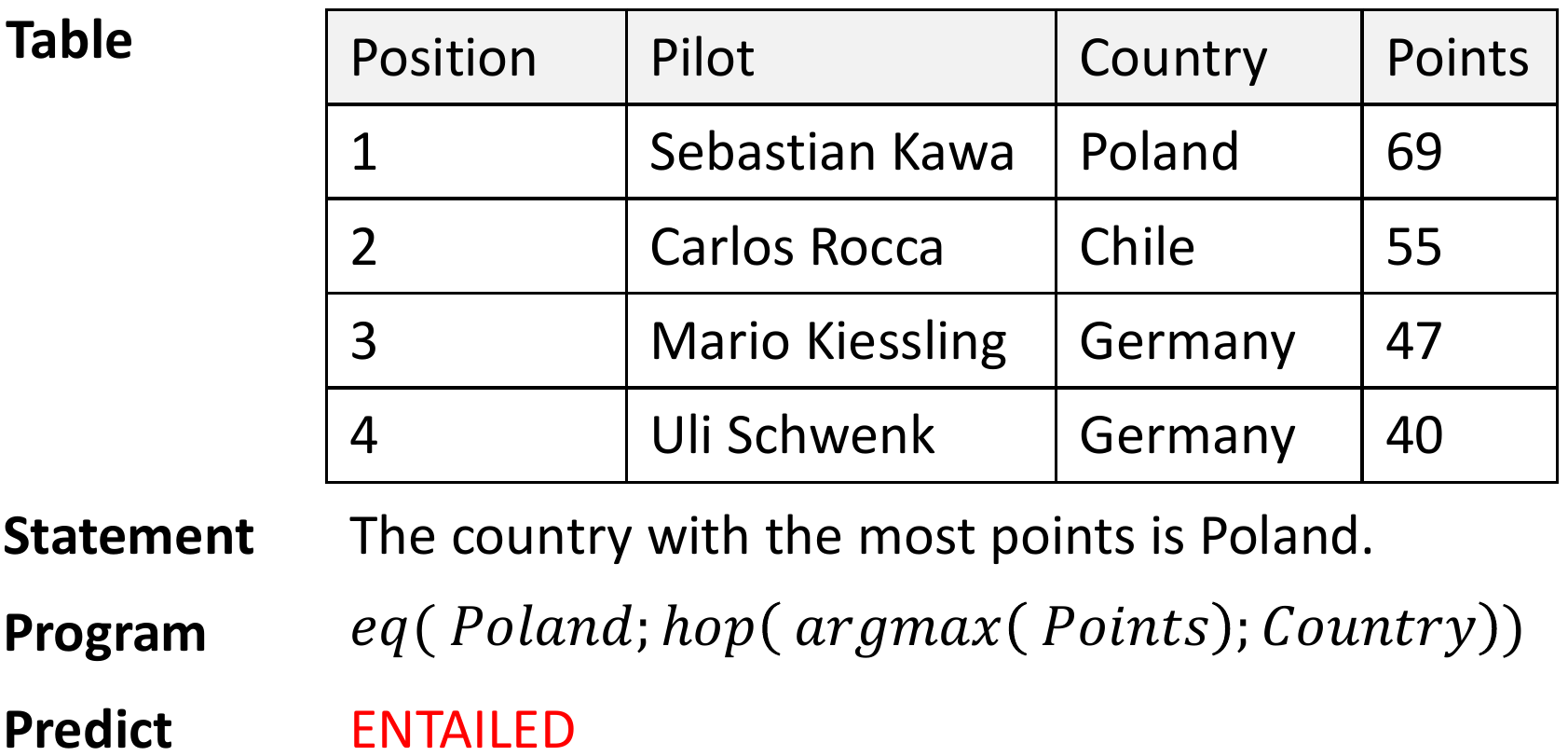}
	\caption{A case study of our approach.}
	\label{fig:case-study}
    \end{figure}

    We conduct a case study by giving an example shown in Figure \ref{fig:case-study}.
    From the example, we can see that our system synthesizes a semantic-consistent program of the given statement and make the correct prediction utilizing the synthesized program.
    This observation reflects that our system has the ability to 
%    (1) match the semantic meaning between the synonyms in semantic parsing phase 
    (1) find a mapping from the textual cues to a complex function (such as the mapping from ``\textit{most points}" to function ``\textit{argmax}") and
    (2) derive the structure of logical operations to represent the  semantic meaning of the whole statement.
    % The semantic meaning of ``most points" is hard to be captured only by semantic meaning of words. To solve it, our 
    % \modelname first generate semantically correct program by the semantic parser and output the correct answer based on learning the semantic compositionality of the program.
    
    \begin{figure*}[ht]    \centering
	\includegraphics[width=\textwidth]{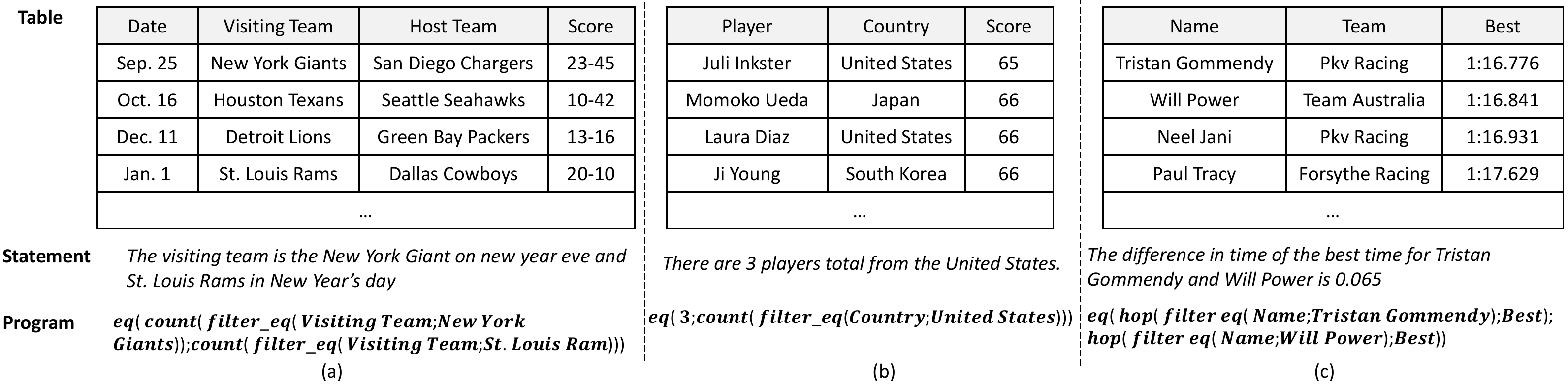}
	\caption{Examples of error types, including (a) predicting a wrong program because of the lack of background knowledge, (b) predicting a correct program but predicting a wrong label, and (c) that the logical operations required to understand the statement is not covered in the grammar.}
	\label{fig:error-1}
    \end{figure*}
    
	\subsection{Error Analysis}
	We randomly select 400 instances and summarize the major types of errors, which can be considered as future directions for further study.
	\par 
    % The error types can be divided into two situations: 
    % (1) The generated programs have incorrect semantics, (2) The generated programs are correc 
	The \textbf{dominant} type of errors is caused by the misleading programs generated by the semantic parser. As shown in the example in Figure \ref{fig:error-1} (a), the semantic parser fails to generate a semantically correct program because it lacks the external knowledge about the date in the table and the ``new year eve'' in the statement.
% 	``lose a game" is related to the comparison of score.
% 	\par
	The \textbf{second} type of errors is caused by semantic compositionality, even though programs are correctly predicted. As shown in Figure \ref{fig:error-1} (b), the program involves operations requiring complex reasoning, like counting the exact number of rows.
	Potential ways to alleviate this problem is to design more function-specific modules like \newcite{Andreas2015DeepCQ}.
% 	Failing to model the synonym in entity-linking step causes the second type of errors. Entity-linking step is essential because program generation requires linking entities in statement to the columns names of tables. For example, a statement claims a fact about ``racers" but the corresponding column name is ``rider", which may lead to the failure of program generation. 
% 	\begin{figure}[h]    \centering
% 	\includegraphics[width=0.5\textwidth]{error-analyze-2.pdf}
% 	\caption{An example for the third error type.}
% 	\label{fig:error-2}
%     \end{figure}
% 	The third type of errors is caused by the challenge of dealing with counting problems. As shown in the Figure \ref{fig:error-2}, solving these problems always require more precise reasoning over a subview of the whole table. 
	The \textbf{third} type of errors is caused by the coverage of the logical operations we used. In this work, we follow \citet{2019TabFactA} and use exactly the same functions. However, as shown in \ref{fig:error-1} (c), understanding this statement requires the function of \textit{difference\_time}, which is not covered by the current set.
	\section{Related Work}
	There is a growing interest in fact checking 
% 	draws increasing attention 
	in NLP with the rising importance of 
	% 	an important task in natural language processing, with the goal of automatically 
	assessing the truthfulness of texts, especially when pre-trained language models \cite{radford2019language,zellers2019defending,keskar2019ctrl} are more and more powerful in generating fluent and coherent texts. 
	Previous studies in the field of fact checking differ in the genres of supporting evidence used for verification, including natural language \cite{thorne2018fever}, semi-structured tables \cite{2019TabFactA}, and images \cite{zlatkova2019fact,nakamura2019r}.
	
	The majority of previous works deal with textual evidence. 
	FEVER \cite{thorne2018fever} is one of the most influential datasets in this direction, where evidence sentences come from 5.4 million Wikipedia documents. 
	Systems developed on FEVER are dominated by pipelined approaches with three separately trained models, i.e. document retrieval, evidence sentence selection, and claim verification.
	There also exist approaches \cite{yin2018twowingos} that attempt to jointly learn evidence selection and claim verification.
	More recently, the second FEVER challenge \cite{thorne-etal-2019-fever2} is built for studying adversarial attacks in fact checking\footnote{\url{http://fever.ai/}}.
	Our work also relates to fake news detection. For example,  \newcite{rashkin2017truth} study fact checking by considering  stylistic lexicons, and \newcite{wang2017liar} builds LIAR dataset with six fine-grained labels and further uses meta-data features.
	There is a fake news detection challenge\footnote{\url{https://www.kaggle.com/c/fake-news-pair-classification-challenge/}} hosted in WSDM 2019, with the goal of the measuring the truthfulness of a new article against a collection of existing fake news articles before being published.
	There are very recent works on assessing the factual accuracy of the generated summary in 
% 	The WSDM 2019 Fake News Classification 
	% 		Fake news detection \cite{wang2017liar,rashkin2017truth}
	neural abstractive summarization systems \cite{goodrich2019assessing,kryscinski2019evaluating}, as well as the use of this factual accuracy as a reward to improve abstractive summarization \cite{zhang2019optimizing}.
% 	\par

 	\citet{2019TabFactA} recently release TABFACT, a large dataset for table-based fact checking.
 	Along with releasing the great dataset, they provide two baselines: Table-BERT and LPA.
% % 	propose two strong baselines for table-based fact checking. 
 	Table-BERT is a textual matching based approach, which takes the linearized table and statement as inputs and states the veracity.
 	However, Table-BERT fails to utilize logical operations.
 	% % 	which tackle the problem as a natural language inference (NLI) problem by linearizing a table as a sequential premise sentence and stating whether the statement is entailed by the table.
% % 	Although this method can capture the semantic meaning of words,
% 	However, it fails to capture the rich symbolic information embodied in program and 
% % 	on reasoning over the inner structure of contexts and 
% 	program-guided semantic compositionality.
% % 	of logical operations.
% 	%The first is that these methods lose the structural information of the table, which make they hard to do deep reasoning over the table.
 	LPA is a semantic parsing based approach, which first synthesizes
 	programs by latent program search and then ranks candidate programs with a neural-based discriminator.
 	However, the ranking step in LPA does not consider the table information.
% 	Since LPA is built based on human-defined lexicons, it may 
% % 	be relatively weak in modeling semantic meaning of words in the statement and 
% 	have lower coverage of semantic-consistent programs.
 	Our approach simultaneously utilizes the logical operations for semantic compositionality and the connections among tables, programs, and statements. Results show that our approach 
% 	leverages the advantages of both, and 
achieves the state-of-the-art performance on TABFACT.
% 	innovative neural model
% 	which constructs LISP-like program candidates based on entity-linking techniques and human-defined rules. 
% 	LPA further selects the most consistent program by a trained discriminator.
% 	Moreover, LPA selects the most consistent program by training a discriminator that treats programs as sequential sentences and matches them with the given statement, which ignores the structure of given tables and the hierarchical information of programs. 
% 	Although LPA considers logical operations for fact checking, the coverage of LPA might be lower than neural-based methods. 
% 	To address the aforementioned issues, we simultaneously consider both the semantic meaning of words, structure of the inputs and semantic compositionality of logical operations, which is an essential direction towards complex reasoning for fact checking.
	
	\section{Conclusion}
	In this paper, we present \modelname, a neural network based approach that considers logical operations for fact checking. 
	We evaluate our system on TABFACT, a large-scale benchmark dataset for verifying textual statements over semi-structured tables, and demonstrate that our approach achieves the state-of-the-art performance.
	\modelname\ has a sequence-to-action semantic parser for generating programs, and builds a heterogeneous graph to capture the connections among statements, tables, and programs.
	We utilize the graph information with two mechanisms, including a mechanism to learn graph-enhanced contextual representations of tokens with graph-based attention mask matrix, and a neural module network which learns semantic compositionality in a bottom-up manner with a fixed set of modules.
% 	along the graph structure 
	We find that both graph-based mechanisms are beneficial to the performance, and our sequence-to-action semantic parser is capable of generating semantic-consistent programs.
\section*{Acknowledge}
Wanjun Zhong, Jiahai Wang and Jian Yin are supported by the National Natural Science Foundation of China (U1711262, U1611264,U1711261,U1811261,U1811264, U1911203), National Key R\&D Program of China (2018YFB1004404), Guangdong Basic and Applied Basic Research Foundation (2019B1515130001),  Key R\&D Program of Guangdong Province (2018B010107005).  The corresponding author is Jian Yin.
	
% 	Our approach not only conduct semantic reasoning over the meaning of words in the contexts, but also consider the inner structure of contexts and connection between them by building a heterogeneous graph. Moreover, our approach also utilize logical operations by learning its semantic compositionality by neural module networks. Experiments show that both graph-enhanced contextual learning module and semantic compositionality learning module bring improvement on the performance and our system achieves state-of-the-art result on TABFACT, a benchmark dataset on table-based fact checking.
% 	\par
% 	In the future, we plan to improve our system by designing more complex module for complex reasoning problem such as counting problem. We also plan to improve the semantic parser to capture the more accurate semantic meaning of statements.
	\bibliography{acl2020}
	\bibliographystyle{acl_natbib}
	\appendix
	\section{Statistic of TABFACT}
	\label{appendix:tabfact}
	\begin{table}[thb]
		\centering
		\small
		\begin{tabular}{ccccc} 
			\toprule
			%\multicolumn{1}{l}{\multirow{2}{*}{Channel}} & \multicolumn{1}{c}{\multirow{2}{*}{\#Sentence}} & \multicolumn{1}{c}{\multirow{2}{*}{\#Table}} & \multicolumn{2}{c|}{Length} & \multirow{2}{*}{Split}  &  \multirow{2}{*}{\#Sentence} & \multirow{2}{*}{Table} & \multirow{2}{*}{Row} & \multirow{2}{*}{Col} \\ 
			%\multicolumn{1}{l}{} & \multicolumn{1}{l}{}                        & \multicolumn{1}{l}{} & Entailed  & Refuted & & & & &\\ 
			Split & \#Sentence &  Table & Avg. Row & Avg. Col\\
			\midrule
			Train & 92,283 & 13,182 & 14.1 & 5.5       \\
			Val & 12,792 & 1,696 & 14.0 & 5.4        \\
			Test & 12,779 & 1,695 & 14.2 & 5.4        \\
			\bottomrule
		\end{tabular}
		\caption{Basic statistics of Train/Val/Test split in the dataset}
		\label{tab:train_val_test}
	\end{table}
	
	\section{Training Details}
	\label{appendix:training-detail}
	In this part, we describe the
	training details of our experiments. As described before, the semantic parser and statement verification model are trained separately.

	We first introduce the training process of the semantic parser. Both training and validation datasets are created in a same way as described in \cref{sec:program-generation}.
	Specifically, each pair of data is labeled as true or false. Finally, the training dataset contains 495,131 data pairs, and the validation dataset contains 73,792 data pairs.
	We implement the approach with the LSTM-based recurrent network and use the following set of hyper parameters to train models: hidden size is 256, learning rate is 0.001, learning rate decay is 0.5, dropout is 0.3, batch size is 150. We use glove embedding to initialize embedding and use Adam to update the parameters. We use beam search during inference and set beam size as 15. We use BLEU to select the best checkpoint by validation scores.

	Then we introduce the training details of statement verification model.
	We employ cross-entropy loss as the loss function. We apply AdamW as the optimizer for model training. In order to directly compare with Table-BERT, we also employ BERT-Base as the backbone of our approach. 
	The BERT network and neural module network are trained jointly. We set learning rate as 1e-5, batch size as 8 and set max sequence length as 512. The training time for one epoch is 1.2 hours by 4 P40 GPUs. We set the dimension of entity node representation as 200. 
	\section{ASDL-Grammar}
	\label{appendix:asdl-grammar}
	In this part, we introduce the ASDL grammar \cite{yin2018tranx} we apply for synthesizing the programs in Seq2Action model. The definition of functions mainly follows \citet{2019TabFactA}. Details can be found in following two pages.\footnote{The function ``\textit{filter\_eq}" contains three arguments (sub\-table, column\_name, value), but we ignore the first argument in the running example for a clearer illustration.}
	\begin{figure*}[t]
		\includegraphics[width=\textwidth]{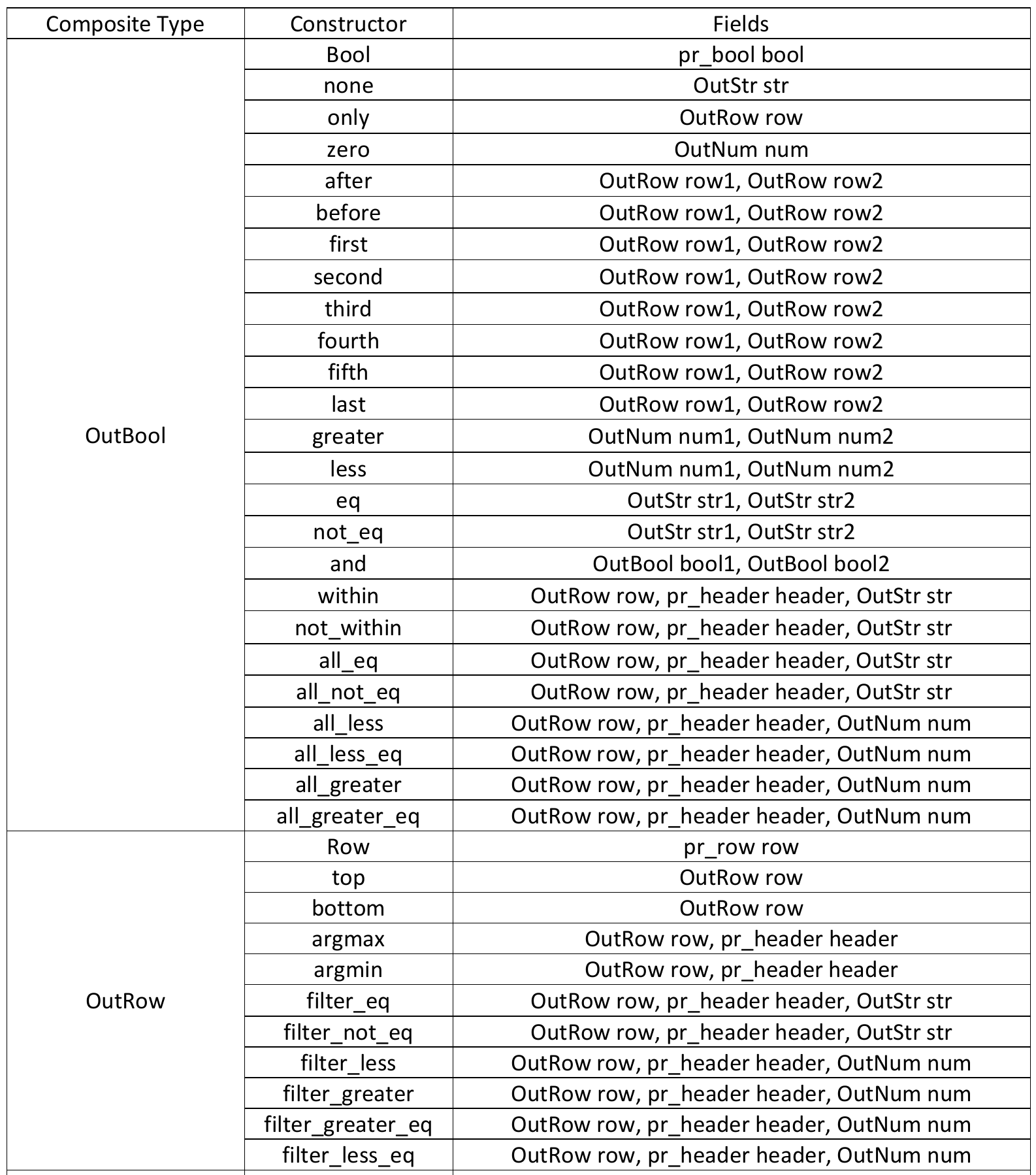}
	\end{figure*}
	\begin{figure*}[t]    
		\includegraphics[width=\textwidth]{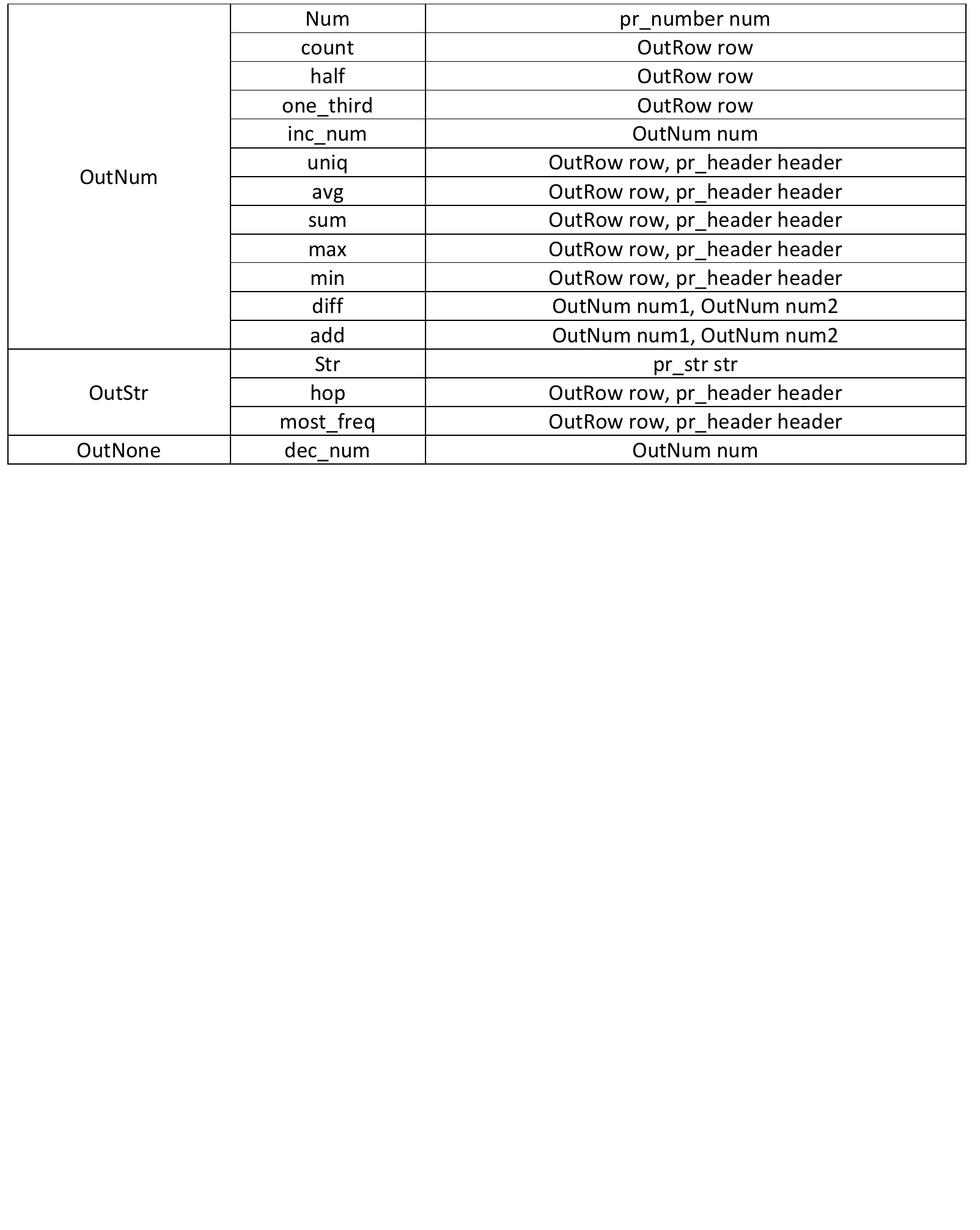}
	\end{figure*}

\end{document}